\pdfoutput=1
\documentclass[10pt]{article}

\usepackage{amsmath}
\usepackage{amssymb}

\usepackage{graphicx}

\usepackage{cite}

\usepackage{color} 

\usepackage[labelfont=bf,labelsep=period,justification=raggedright]{caption}

\makeatletter
\renewcommand{\@biblabel}[1]{\quad#1.}
\makeatother

\date{}

\begin{document}

% Title must be 150 characters or less
\begin{flushleft}
{\Large
\textbf{Evolvability Is Inevitable: Increasing Evolvability Without the Pressure to Adapt}
}
% Insert Author names, affiliations and corresponding author email.
\\
Joel Lehman$^{1,\ast}$, 
Kenneth O. Stanley$^{2}$ 
\\
\bf{1} Department of Computer Science, The University of Texas at Austin, Austin, TX, USA
\\
\bf{2} Department of Electrical Engineering and Computer Science, University of Central Florida, Orlando, FL, USA
\\ 
$\ast$ E-mail: joel@cs.utexas.edu 
\end{flushleft}

% Please keep the abstract between 250 and 300 words
\section*{Abstract}
Why evolvability appears to have increased over
evolutionary time is an important unresolved biological question.
Unlike most candidate explanations, this
paper proposes that increasing evolvability can result without %we
any pressure to adapt.
The insight is that if evolvability is heritable,
then an unbiased drifting process
across genotypes %we
can still create a distribution of phenotypes %we
biased
towards evolvability, because evolvable organisms
diffuse more quickly through the space of possible
phenotypes.
Furthermore, because phenotypic divergence often correlates with founding
niches, niche founders may on average be more evolvable,
which through
population growth provides a genotypic
bias towards evolvability. Interestingly, the combination
of these two mechanisms can lead to increasing evolvability
without any pressure to out-compete other organisms,
as demonstrated through experiments with
a series of simulated models.
Thus rather than from pressure to adapt, evolvability may
inevitably result from any drift through genotypic space
combined with evolution's passive tendency to accumulate niches.
% Please keep the Author Summary between 150 and 200 words
% Use first person. PLoS ONE authors please skip this step. 
% Author Summary not valid for PLoS ONE submissions.   
%\section*{Author Summary}

\section*{Introduction}

An unbroken hereditary chain links 
the simplest early replicators to the most complex 
modern macroscopic organisms. Observing this evolutionary
trajectory raises the
question of the \emph{cause} for the appearance
of increasing evolutionary potential, i.e.\ increasing \emph{evolvability} \cite{dawkins:evolution}. 
Although the cause of such increase is still debated, 
most candidate explanations
for evolvability rely on selection pressure 
\cite{dawkins:evolution,earl:evolvability,kirschner:evolvability,wagner:evolvability,pigliucci:evolvability,brookfield:evolvability,wagner:robustness,bloom:protein}, reflecting natural
selection's significant explanatory power in other contexts.
For example, selection on mutation or recombination 
rates \cite{earl:evolvability}, species-level selection to adapt \cite{dawkins:evolution,kirschner:evolvability}, selection for stability of evolved structures \cite{bloom:protein,kirschner:evolvability}, and
persisting through 
fluctuating selective environments \cite{pigliucci:evolvability} have all 
been proposed as partial explanations for increasing evolvability.
However, adaptive explanations may be unnecessary or at least
merit more scrutiny if
increasing evolvability is demonstrated 
\emph{without} any pressure to adapt, that is, if evolvability
results from a more fundamental (and potentially passive) process.

This paper investigates two such alternative hypotheses for evolvability.
The first hypothesis is that if evolvability itself is heritable, then even
a passive drifting process over genotypes
will differentiate the
evolvability of organisms, and the more evolvable of these organisms 
will be more likely to become phenotypically diverse and spread
through niches. That is, a biased
distribution of phenotypes can result from a passive drift over genotypes.

Intuitively, in a passive drift some mutations may increase an organism's 
evolutionary potential, while
others may decrease it. 
Importantly, note that such passive drifting does not cause
an inherent drive towards increasing evolvability when averaged over
all genotypes in the entire population.
However, it turns out that 
evolvability averaged \emph{over niches}  
may still rise even in a purely drifting model (i.e.\ a model with a 
fixed-size population that evolves solely through genetic drift).
Following one widely-held conception of evolvability
\cite{brookfield:evolvability,wagner:evolvability,dichtel:evolvability,kirschner:evolvability}, those organisms 
that are least evolvable will on average change less
phenotypically from repeated mutation, while those that are more evolvable
will change more, i.e.\ more evolvable organisms will have a higher
average 
\emph{velocity} of phenotypic change. 

As a result, the phenotype space itself can
act as a filter, whereby more evolvable organisms will
be separated from the less evolvable over time as they
 radiate at different velocities throughout the phenotypic space. 
This sorting mechanism is similar to how a centrifuge or a western blot 
separates particles of different densities or charges.
In other words, at any point in time 
the least evolvable organisms are most likely to
be found clustered together within the phenotypic space,
occupying niches near their evolutionary origins.
In contrast, the more evolvable organisms are more likely to diverge
phenotypically over time to inhabit
niches divergent from their ancestors.
Thus, even if the genotypic space is evolving
without direction, the resulting distribution
in the phenotypic space can still 
 become biased towards the more evolvable.
That is, uniformly sampling the genotype space (which is unbiased)
would on average choose
less evolvable organisms than would uniformly sampling the phenotype
space (which is biased). The bias in the distribution of phenotypes
is that less evolvable organisms
are likely to be found densely concentrated in only a few niches (near their
evolutionary origin), while the 
more evolvable organisms are more likely to spread 
\emph{throughout} reachable niches.

Thus if a population is drifting through a genotypic space,
from surveying only the phenotypic
space it might be mistakenly inferred 
that the average evolvability over all organisms had \emph{increased},
i.e.\ that there is a genotypic bias towards evolvability. Furthermore,
the cause of this apparent increase might be misattributed to selection
pressure. 
In reality, however, there is no selection pressure, and 
the average evolvability of genotypes will 
not have significantly changed: Only the average evolvability per
niche (i.e.\ averaged over divergent phenotypes) will have increased. 
The interesting implication is
that the deceptive appearance of
increasing evolvability can result from a random walk over genotypes. 
However, the main insight is
that evolvability may be self-reinforcing: A drifting process in
the genotypic space may warp the phenotypic distribution in proportion to
evolvability, and given a sufficiently large population, 
the \emph{maximum} evolvability may
also increase over time, which further warps the phenotypic space.
Supporting this hypothesis, experiments with both an abstract mathematical
model and simulated evolved machines reveal the appearance of increasing 
evolvability through only a drifting process.

However, 
while genetic drift biases only the phenotypic
distribution of organisms towards greater evolvability, 
an additional non-adaptive mechanism may also similarly bias the
\emph{genotypic} distribution.
This genotypic bias can result from the correlation between
 phenotypic divergence and establishing new niches. In other words,
evolvable organisms may be more likely to lead to new ways of life \cite{kirschner:evolvability}. 
Thus more evolvable organisms may become over-represented
as founders of new niches, causing the resulting population growth
from niche foundation to bias the genetic space also towards increasing
evolvability. Thus the second hypothesis for non-adaptive evolvability
increase is that founder effects in new niches 
tend to amplify more evolvable organisms on average.
The end result is that \emph{overall} evolvability, i.e.\ not just its
appearance, may also
 increase over time in nature -- but not due to adaptive pressure to 
out-compete other organisms, which is a foundational assumption that
underlies many other theories for the rise of evolvability
\cite{dawkins:evolution,earl:evolvability,kirschner:evolvability,wagner:evolvability,pigliucci:evolvability,brookfield:evolvability,wagner:robustness,bloom:protein}.

Supporting this second hypothesis, 
further experiments with growing populations 
in which evolution is initiated within
a single niche, and where each niche has a limited capacity
(but where selection is random within a niche) demonstrate a significant
trend towards increasing \emph{genotypic} 
evolvability over time. Importantly, the drive towards overall
increasing genotypic evolvability in these experiments is qualitatively 
more substantial than
 in the drifting models alone (where the appearance of 
increasing evolvability results only
when averaged over niches). Another abstract model and
two additional models with evolved machines exhibit the same
trend towards increasing genotypic 
evolvability without selection pressure for
adaptation. 
The surprising conclusion is that increasing evolvability may 
not result from selective pressure to adapt, but may instead be an inevitable
byproduct of how evolvability warps the distribution of phenotypes
 and the tendency for founding new niches to amplify evolvable organisms.

\section*{Appearance of Increasing Evolvability in Passive Drift Models}

The first set of experiments illustrate that evolvability can appear
to increase as a result of a passive drifting process over genotypes.
That is, if 
evolvability is heritable then a drifting process in a genotype space
can separate the more evolvable organisms from the less evolvable
ones over time, inducing a distorted distribution in phenotype space that 
yields the deceptive impression of overall increasing evolvability. 
This first hypothesis is explored in two models,
a highly-abstract model and a model based on simulated evolved robots.

\subsection*{Abstract Passive Drift Model}

The highly-abstract model consists of
a population of abstract organisms that evolve solely due to 
genetic drift (i.e.\ there is no selection pressure nor population growth). 
The idea
is to investigate whether genetic drift can
yield the appearance of increasing evolvability in a minimal
model.
Thus each organism in this model has only two hereditary
properties: the niche that it occupies and its evolvability, both of
which are subject to mutation. An organism's niche is
represented as a two-dimensional point within a discrete grid, which 
mutation perturbs by shifting the point one unit in either dimension. 
In other words, the genotype-to-phenotype map is trivial in this model:
The niche specified in an organism's genotype maps directly into its
phenotypic niche (which is the two-dimensional point in the discrete grid).
The evolvability of an organism is thus specified
as the probability that an organism's niche will be perturbed through
mutation, which reflects the assumption
that more evolvable organisms have greater phenotypic
variability \cite{brookfield:evolvability,wagner:evolvability,dichtel:evolvability,kirschner:evolvability}.
In contrast to the initial probability for an organism's 
niche to be perturbed (i.e.\ the
organism's \emph{initial} evolvability),
evolvability itself mutates more infrequently through small perturbations
(exact parameter settings can be found in the Methods Section).
In other words, the assumption is that evolvability tends to
evolve at slower rates than typical hereditary properties.
Note that all organisms are initially identical, i.e.
they begin in the same 
niche (in the center of the grid) with the same level of evolvability.

The idea is that as the population drifts in this model, the evolvability
 of each organism undergoes a random walk. 
Thus over time the evolvability of organisms
will become differentiated as by chance mutation some become more evolvable 
and some become less so. 
 However, across the entire population, evolvability
will remain constant on average because increasing and decreasing are 
equiprobable. 
Concurrent with changes in evolvability, the
niches of organisms are also evolving stochastically. 
Recall that in this model
the probability of an organism's niche mutating is linked to its evolvability.

The interesting effect of such linkage is that
it causes the niche space to act as a filter that 
over time 
separates the less evolvable organisms from the more evolvable. In other words,
if all organisms are initialized to start from the same niche, on average
the organisms that become most evolvable (by chance mutation) will also 
evolve to be farthest in the niche space from the starting location (because
evolvability here correlates with an increased future 
chance of changing niches). That is, the most evolvable organisms have
a higher phenotypic velocity of change. As
a result, the least evolvable organisms will on average cluster near the initial niche, and
the most evolvable organisms are more likely to be found along the peripheral 
niches. Thus by
observing the distribution of evolvability across the 
\emph{niche space} (which is equivalent to the phenotypic space in this
simplified model) 
one might falsely conclude that evolvability 
in general has increased. In other words, the average evolvability
\emph{per niche} will have increased.

\begin{figure}
\begin{center}
\includegraphics[height=2in]{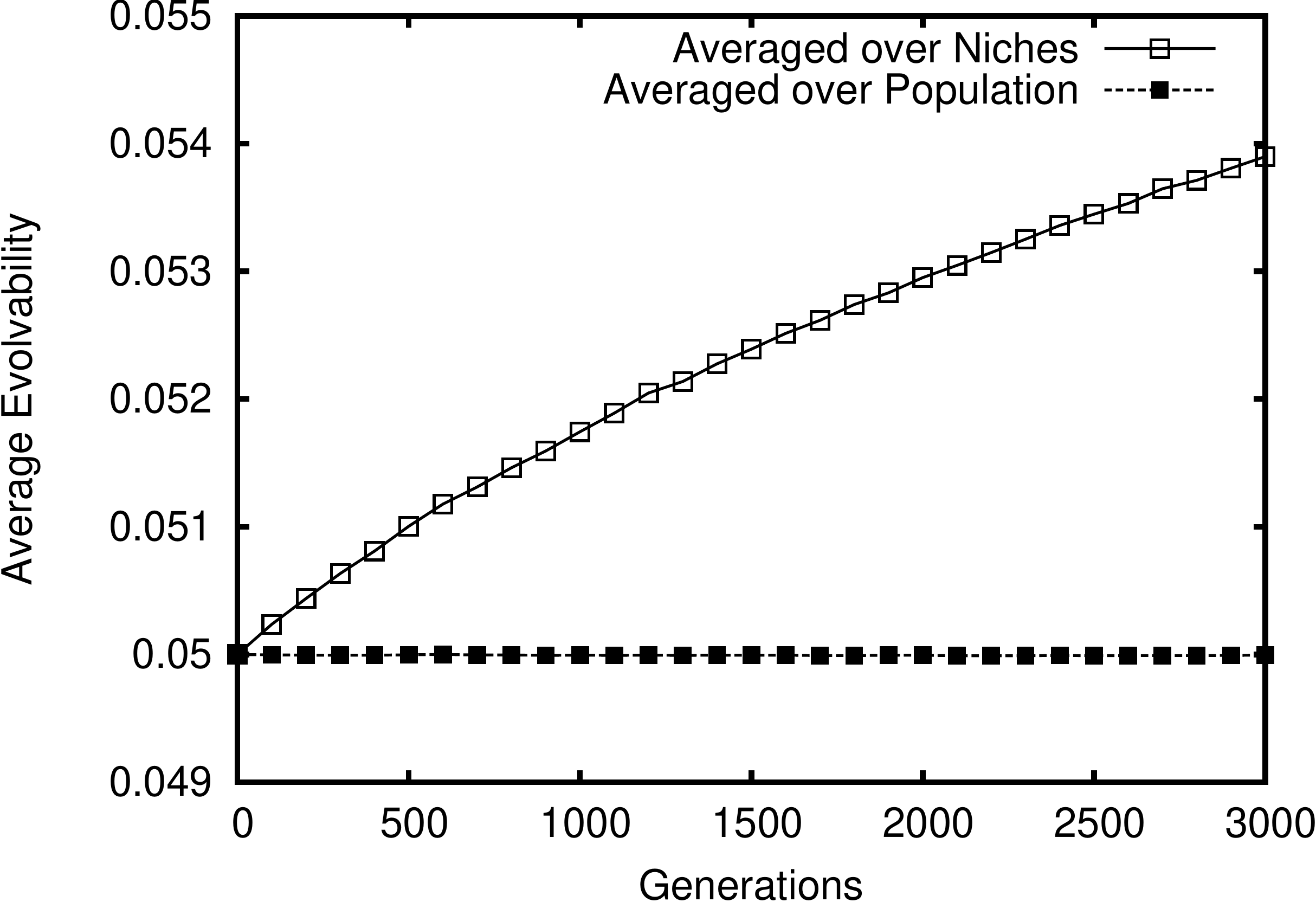}
\caption{\label{fig:model1_evo} \textbf{Evolvability in the abstract passive drift model.} How the evolvability of organisms changes over generations 
is shown averaged (in different ways) over $50$ independent
simulations that last $3,000$ generations each. 
If evolvability is averaged \emph{within} each niche and then \emph{over} all
niches, then evolvability \emph{appears} to increase. However, 
if instead evolvability is simply averaged over all organisms in the population, 
there is no significant overall 
increase in evolvability over time.}
\end{center}
\end{figure}

\begin{figure}
\begin{center}
\includegraphics[height=2in]{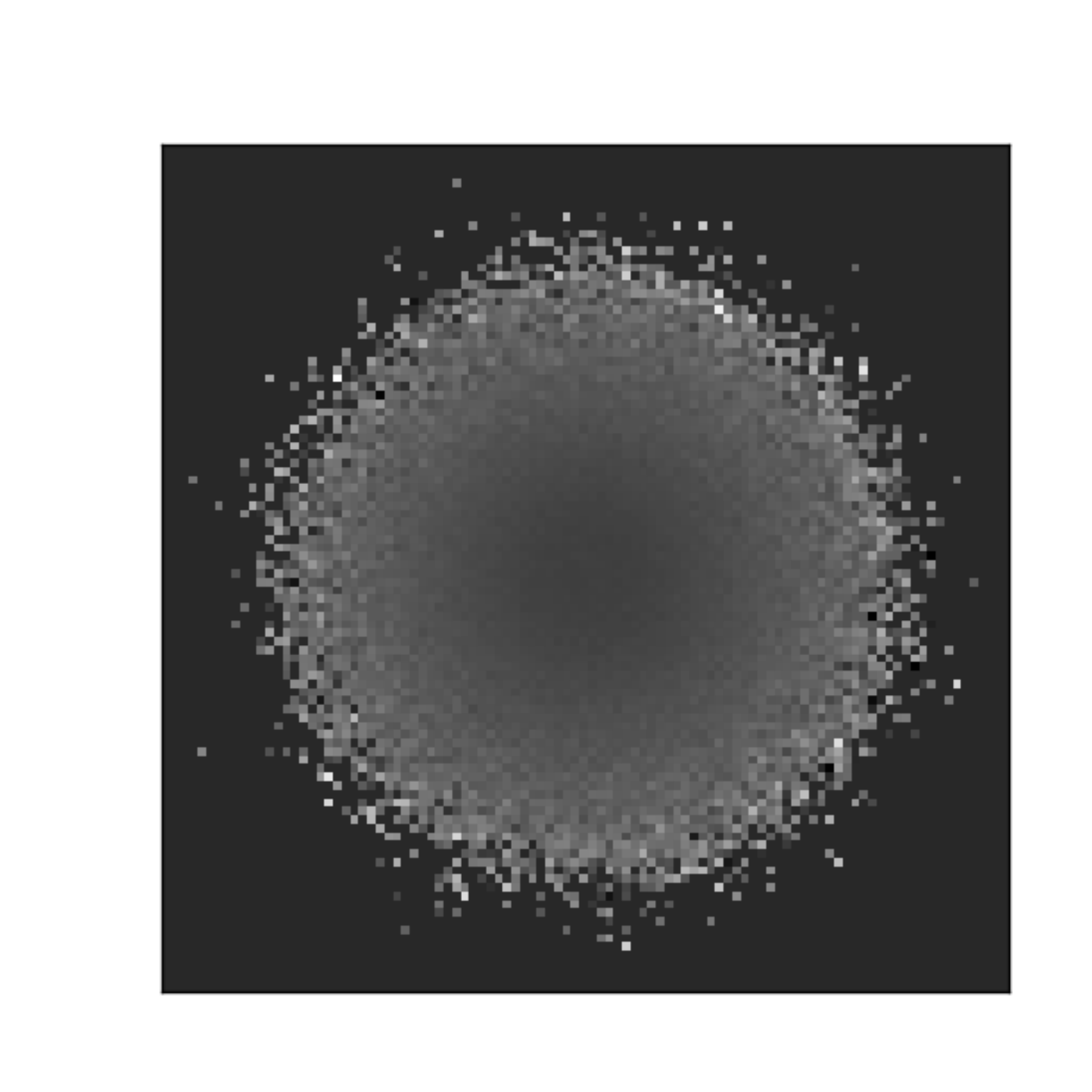}
\caption{\label{fig:model1_niches} \textbf{Evolvability heat map for the abstract passive drift model.} The average evolvability of organisms in each niche
at the end of a simulation is shown averaged over $50$ independent runs.
 The lighter the color, the more evolvable individuals are within that niche. The overall result is that evolvability
increases with increasing distance from the starting niche in the center.
}
\end{center}
\end{figure}

\begin{figure}
\begin{center}
\includegraphics[height=2in]{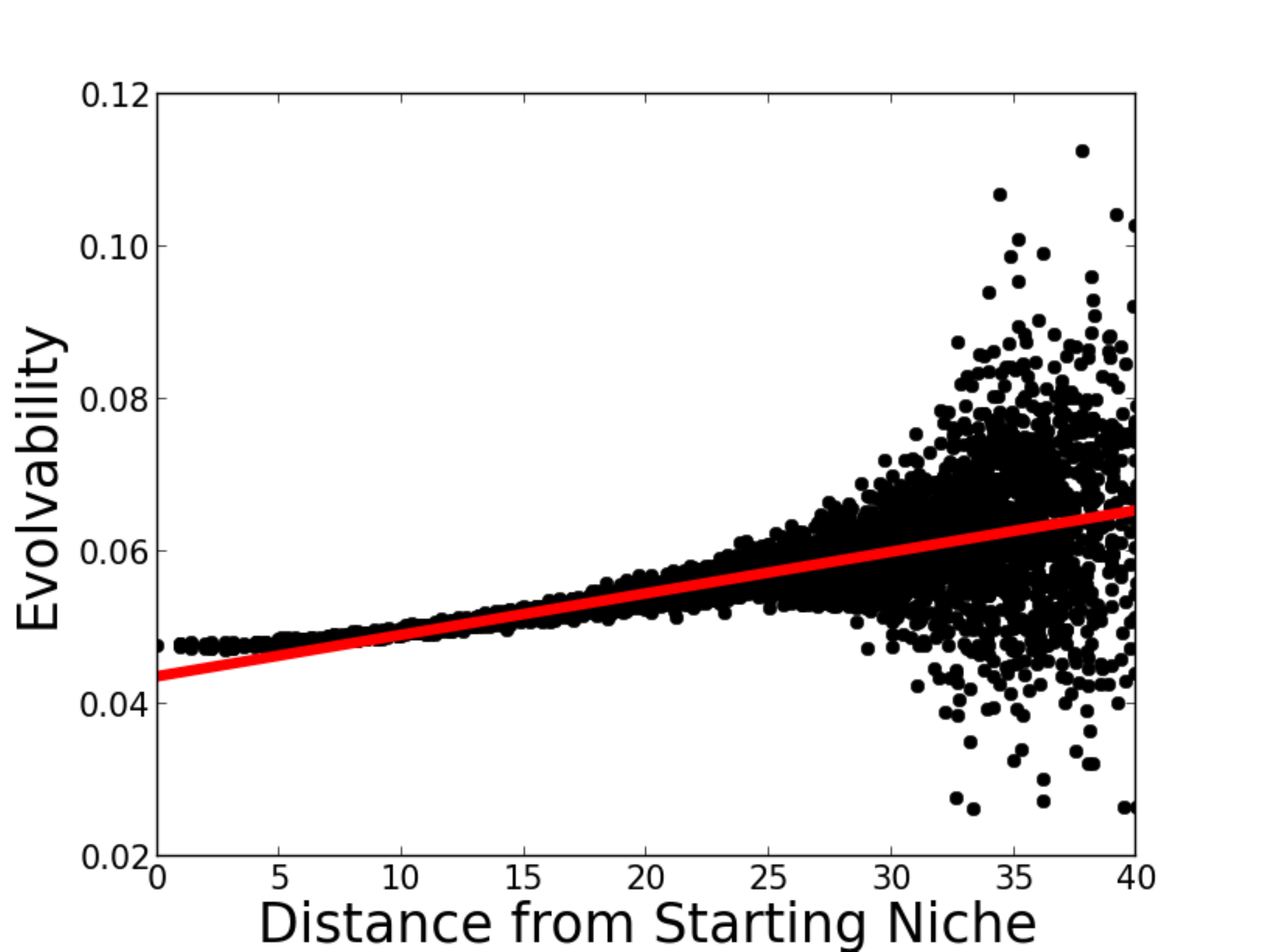}
\caption{\label{fig:model1_distance} \textbf{Evolvability vs.\ distance from
the initial niche for the abstract passive drift model.} 
The evolvability of organisms at the end of a simulation
 is shown as a function of distance from the initial starting
niche (averaged over $50$ independent runs). The main result is that there is
a significant correlation between increasing distance from the initial
niche and increasing 
evolvability in this model.
The plotted line indicates the line of best fit by linear regression.
}
\end{center}
\end{figure}

However, evolvability across the \emph{population}
remains unchanged on average; it is only 
evolvability's distribution over the space of \emph{niches}
 that becomes biased during evolution. This bias, and the unbiased population-wide average of evolvability
are shown in figure \ref{fig:model1_evo}. More clearly illustrating the bias over
the niche space, figure \ref{fig:model1_niches} shows a heat-map of evolvability over the grid of niches at the end of a simulation
and figure \ref{fig:model1_distance} %S1
shows how evolvability
varies as a function of a niche's distance from the
starting niche. Note that there is a strong
monotonic relationship between the distance of an organism from the starting
niche at the end of a simulation 
and its evolvability ($r=0.735, p<0.0001$; Pearson's r).

\subsection*{Passive Drift with Evolved Robots \label{sec:concretepassive}}

To augment the evidence provided by the purely-abstract model in 
the previous section, 
a more concrete genetic space is considered in this model, which has
a richer genotype-to-phenotype map. 
The idea is to begin exploring whether the results from the
abstract passive drift model reflect
a general tendency or if they are overly specific to
parameters or assumptions in the simple model. 
As an initial such exploration, a
genotype-phenotype mapping and simulated environment are implemented in
the spirit of digital evolution \cite{lenski:avida} and 
evolutionary robotics (ER; \cite{nolfi:book00}).
In particular, a genotype-phenotype map for
a simulated robot controlled by 
evolved artificial neural networks (ANNs) is 
adapted from prior ER experiments \cite{lehman:ecj11,mouret:noveltyjp,risi:ab10}. 

In this model, genotypes encode ANNs
 that control simple wheeled robots embedded in a simulated maze.
The motivation is to abstract at a high level how evolved neural structures
influence an organism's behavior in its environment.
In other words, a genotype in this model
 ultimately maps to the behavior of a robot
in a simulated maze environment. Though other domains could be applied,
this environment is well-studied \cite{lehman:ecj11,mouret:noveltyjp,risi:ab10}
and offers a non-trivial genotype-phenotype mapping.
Niches in this model are specified by creating a discrete grid over
a space of possible robot \emph{behaviors}, i.e.\ what the robot actually does
in the simulation, as opposed to a characterization of its genotype or of 
the ANN controller itself to which
the genotype maps. So in this model, as in nature, 
the niche space is a many-to-one
mapping from the space of phenotypes, i.e.\ many similar phenotypic 
behaviors will map into the same niche.
The idea is that different classes of behaviors are
what facilitate niches, not simple differentiation of genotypes or
encoded neural structures. Thus a robot is mapped into a niche as a
function of its behavior.

While there there is no overall consensus on how to quantify 
evolvability \cite{pigliucci:evolvability}, a growing body of work
supports that evolvability is related to phenotypic variability 
\cite{brookfield:evolvability,wagner:evolvability,dichtel:evolvability,kirschner:evolvability}.
Thus the evolvability of genotypes in this model (also 
following precedent in ER \cite{lehman:evolvability}), 
is given by quantifying the amount of behavioral variety (i.e.\ the number of different behavioral niches) 
reachable on average by random mutations from a 
particular genotype. In other words, an
evolvable organism is more likely to lead to phenotypic divergence.
Note that in this model evolvability is not directly encoded into the
genotype as in the abstract model, but is a quantifiable emergent product of
the genotype-phenotype map, more closely resembling the situation in biological
evolution. Specific 
parameters and details about the evolvability measure 
are given in the Methods Section. 

The general motivation for this model 
is that passively drifting with a relatively large population (e.g.\ on the
order of millions) composed of this
kind of more concrete genotype 
may exhibit the same appearance of increasing evolvability 
as observed in the abstract passive drift model. A larger
population is necessary in this experiment because most random genotypes in this
more realistic genotypic space represent similarly 
trivial behaviors, i.e.\ most randomly-connected ANNs 
encode no meaningful information, and the 
appearance of increasing evolvability will only emerge when drifting over a
sufficient quantity of differentiated non-trivial behaviors.
Thus to facilitate these experiments in a computationally efficient way, 
a limited genetic space of ANNs that was
tractable to exhaustively characterize and explore was 
enumerated (i.e.\ a discrete subset was considered from a much larger space of ANNs with continuous weights and variable network topologies).
Then each one of the enumerated genotypes in the space (which consisted
of $38.7$ million different genotypes)
were evaluated in the maze environment to quantify
its behavior and evolvability. For each simulation 
the population was uniformly initialized to
a randomly chosen genotype (i.e.\ the population is 
always initially homogeneous)
and subject to differentiating genetic drift for $250$ generations.
Importantly, supporting the assumption that evolvability is heritable,
despite overall wide variance in evolvability over the entire genetic space,
the evolvability of a parent and its offspring are well correlated ($r=0.770$; Pearson's r).
The details of the particular 
neural model applied can also be found in the Methods Section.

\begin{figure}
\begin{center}
\includegraphics[height=2in]{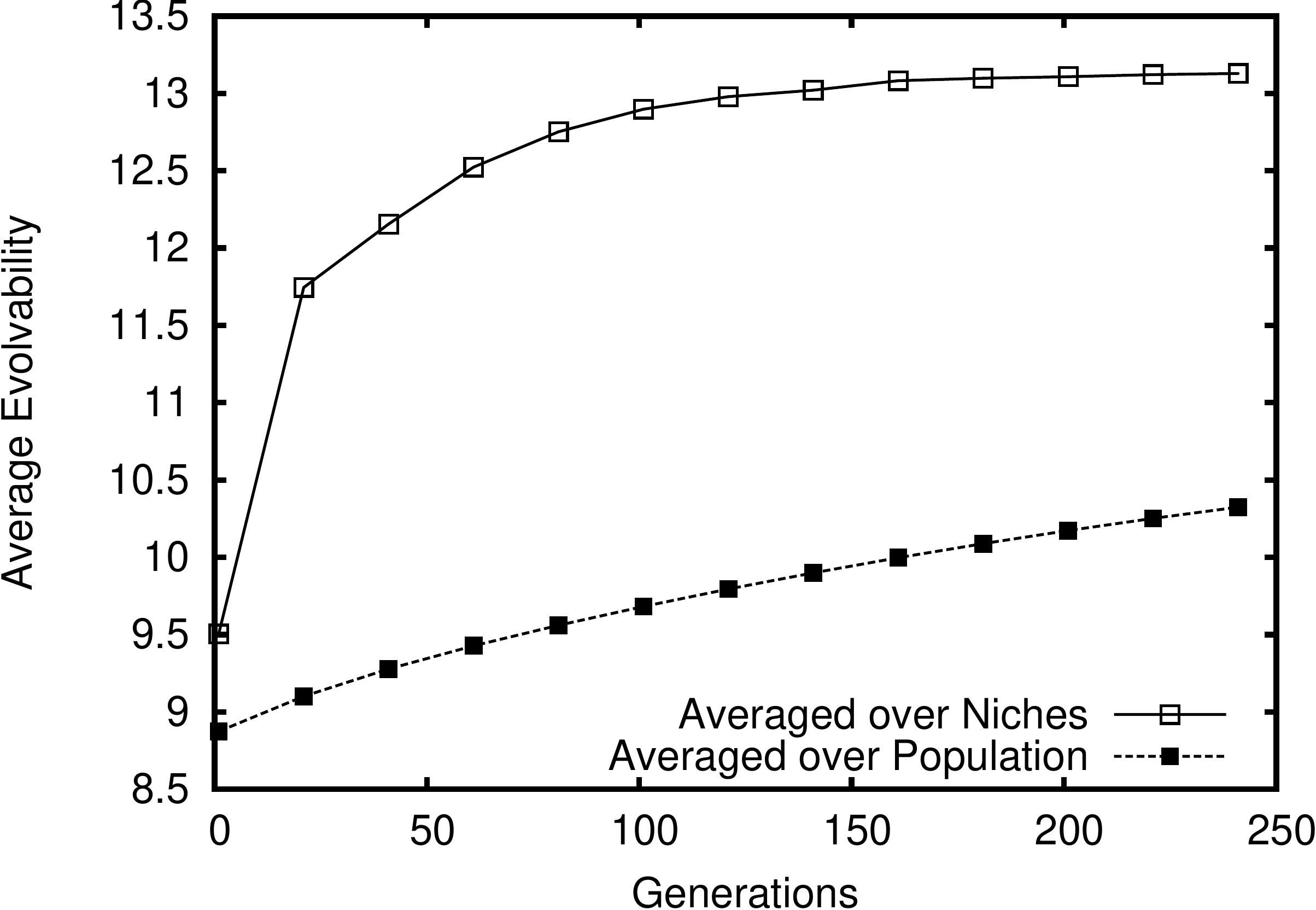}
\caption{\label{fig:model3_evo} \textbf{Evolvability of evolved robots with the passive drift model.} The evolvability of evolved robots subject to
passive drift is shown averaged (in different ways) over $50$ independent
runs that lasted $250$ generations each. If evolvability is averaged within each niche and then over all
niches, it appears to increase. If instead evolvability is averaged over all organisms, there is no
significant increase.}
\end{center}
\vspace{-0.1in}
\end{figure}

Interestingly, a plot of evolvability over time from this model 
also demonstrates the appearance of increasing
evolvability over time (figure \ref{fig:model3_evo}), providing evidence
that the hypothesis that genetic drift can lead to the appearance 
of increasing evolvability
may hold true not only in abstract theoretical circumstances. 

\section*{Population Growth with Limited Capacity Niches Increases Evolvability}

The previous models demonstrate how a purely random drifting
process can create the deceptive appearance of increasing 
overall evolvability.
The next experiments explore the hypothesis that a qualitatively 
more pervasive increase in 
evolvability (i.e.\ an overall bias towards genotypes with higher evolvability)
 can result from population growth and niches with limited capacity. 
First, an extension of the abstract
passive drift model is considered.

\subsection*{Abstract Model with Limited Capacity Niches \label{sec:abstractniched}}

This section considers a variation of the abstract model introduced
earlier, but where the size
of the population varies dynamically (previously this size was fixed). 
In particular, the population 
grows geometrically (i.e.\ each organism is replaced by two offspring in
the next generation), but the overall size of the population remains
 tractable because in this model niches are limited in capacity
(i.e.\ when a new generation is created, niches can grow only to a certain
size, after which further individuals entering that niche are discarded).
The idea is to roughly model the concept of limited resources in
natural evolution and explore its effect on evolvability.
Importantly, this extended model still imposes no \emph{direct} selection
pressure for evolvability, and if the model started at 
equilibrium (i.e.\ with all niches at full capacity), there would be no
expectation of evolvability increasing over time. 
In other words, what is important for evolvability in this model is 
spreading through the space of niches.
Furthermore, selection
\emph{within} a niche is purely random -- there is no selection for adaptation
to the niche nor any way for one organism to reliably \emph{out-compete} another.
Thus this is a model without adaptive pressure.

\begin{figure}
\begin{center}
\includegraphics[height=2in]{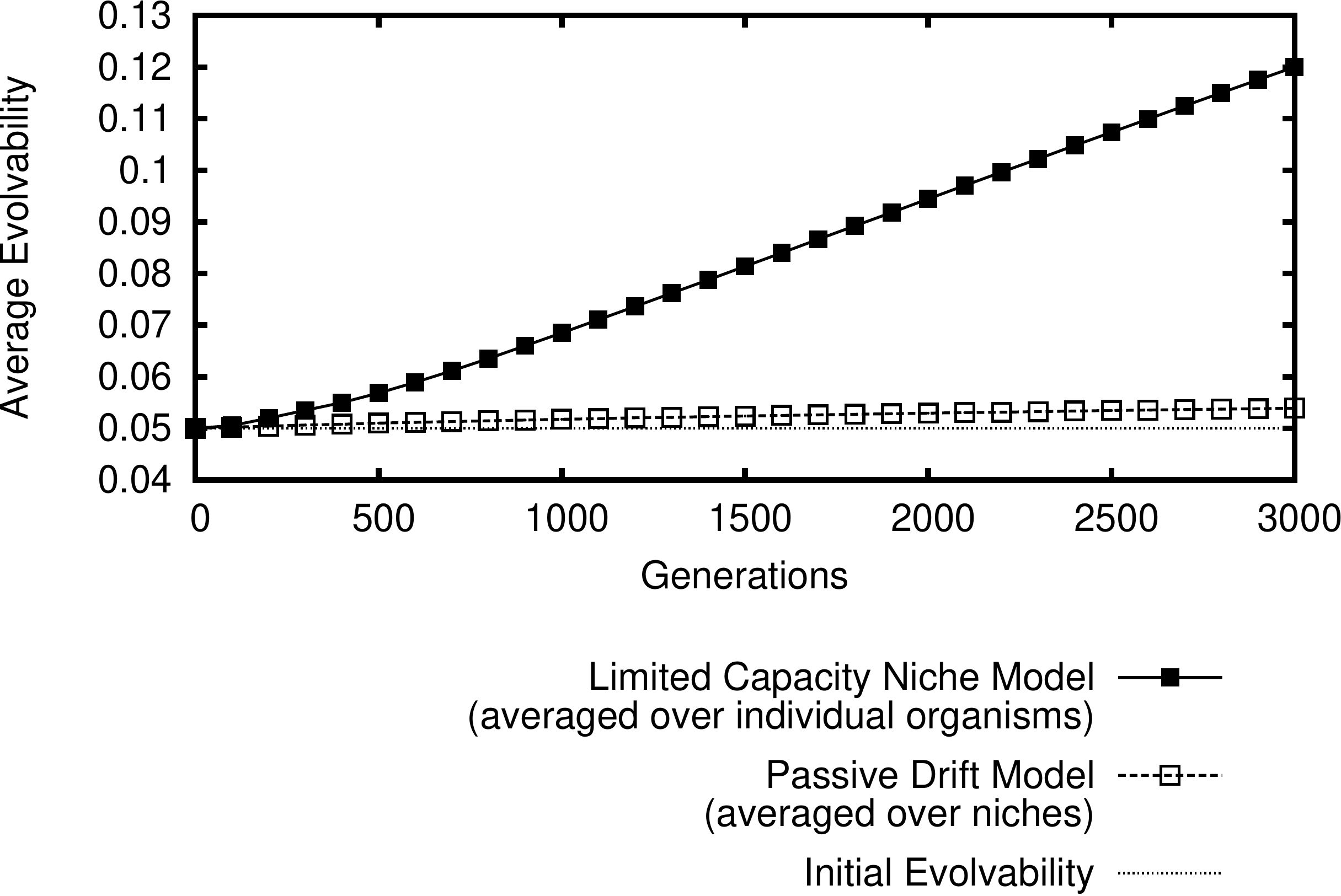}
\caption{\label{fig:model2_evol} \textbf{True increasing evolvability in the abstract model with limited capacity niches.} How the average 
evolvability of organisms in the population changes over time 
is shown (averaged over $50$ independent
simulations 
that lasted $1000$ generations each). Note that the line shown for the
passive model (reproduced from figure \ref{fig:model1_evo}) represents only the \emph{appearance} of increasing evolvability
in that model when evolvability is averaged over niches.}
\end{center}
\end{figure}

However, despite this lack of adaptive pressure,
as evolution progresses in this model
the passive filtering effect of the phenotypic space
demonstrated in the fixed-sized 
population model is amplified.
The explanation is that the resulting population growth 
from founding a new niche (by mutating out of the zone of 
previously explored niches) \emph{indirectly}
rewards increasing evolvability in this model: 
The more
evolvable organisms (which because of their higher velocity of phenotypic
change are more likely to mutate into new niches) are continually 
amplified from population growth as they diffuse through niches.
Thus as more niches are discovered and filled, the
population becomes increasingly biased towards evolvability;
in effect, the reward for discovering a new niche in this model 
accelerates the
filtering process that is purely passive 
in the passive drifting model.
This acceleration is shown in figure \ref{fig:model2_evol}, which compares
evolvability in this model to that of the passive drift model introduced
earlier. Note that
the figure shows growth in overall evolvability (i.e.\ averaged over
all organisms) in the
limited capacity niche model that greatly outpaces the growth in
the passive drift model (which is only significant when averaged over niches).
In other words, superficial niche-level evolvability in the passive
drift model grows more slowly than evolvability over all genotypes in
the limited capacity niche model. This result is important because
it demonstrates a true increase in average evolvability over a population
without selection pressure to out-compete other organisms.

\begin{figure}
\begin{center}
\includegraphics[height=2in]{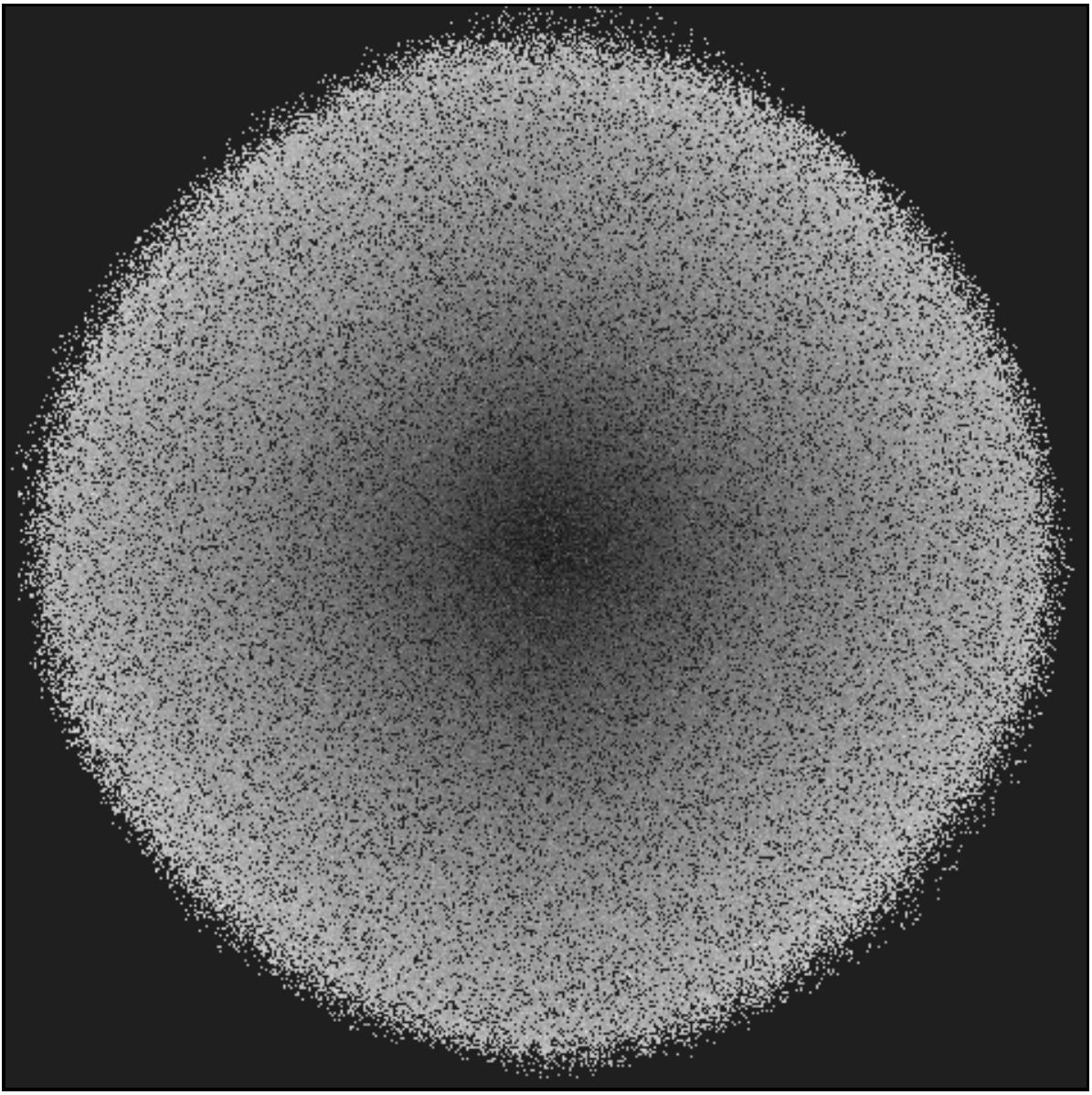}
\caption{\label{fig:model2_niches} \textbf{Evolvability heat map for the abstract model with limited capacity niches.} 
The average evolvability of organisms in each niche
at the end of a simulation is shown. The lighter the color, the more evolvable individuals are within that niche. The overall result is that, as in the
first model, evolvability
increases with increasing distance from the starting niche in the center.
}
\end{center}
\end{figure}

\begin{figure}
\begin{center}
\includegraphics[height=2in]{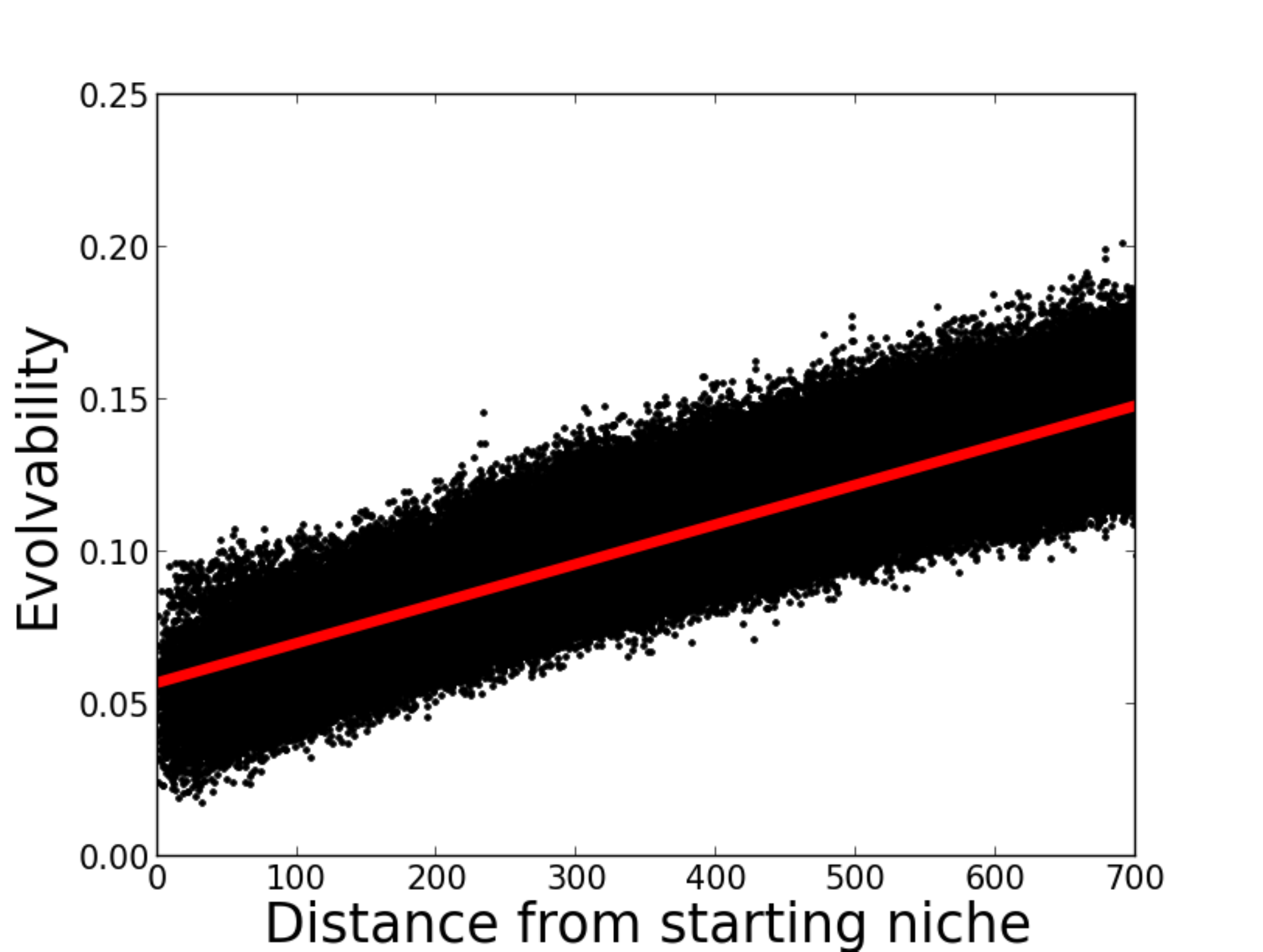}
\caption{\label{fig:model2_distance} \textbf{Evolvability vs.\ distance from
the initial niche for the abstract model with limited-capacity niches.} 
The average evolvability of organisms in the final population is shown as a function of distance from the initial starting
niche averaged over $50$ independent simulations. The main result is that there is
a significant correlation between increasing distance from the initial
niche and increasing 
evolvability. 
The plotted line indicates the line of best fit by linear regression.
}
\end{center}
\end{figure}

Similarly to the first model, figure \ref{fig:model2_niches}  %S2
shows a heat-map of evolvability over the niche space averaged
over all runs, and figure \ref{fig:model2_distance} %S3
shows how evolvability
varies over the niche space as a function of a niche's distance from the
starting niche. There is a strong significant
monotonic relationship between the distance from starting
niche and evolvability ($r=0.967, p<0.0001$; Pearson's r).

\subsection*{Evolved Robots Model with Limited Capacity Niches \label{sec:concreteniche}}

\begin{figure}
\begin{center}
\includegraphics[height=2in]{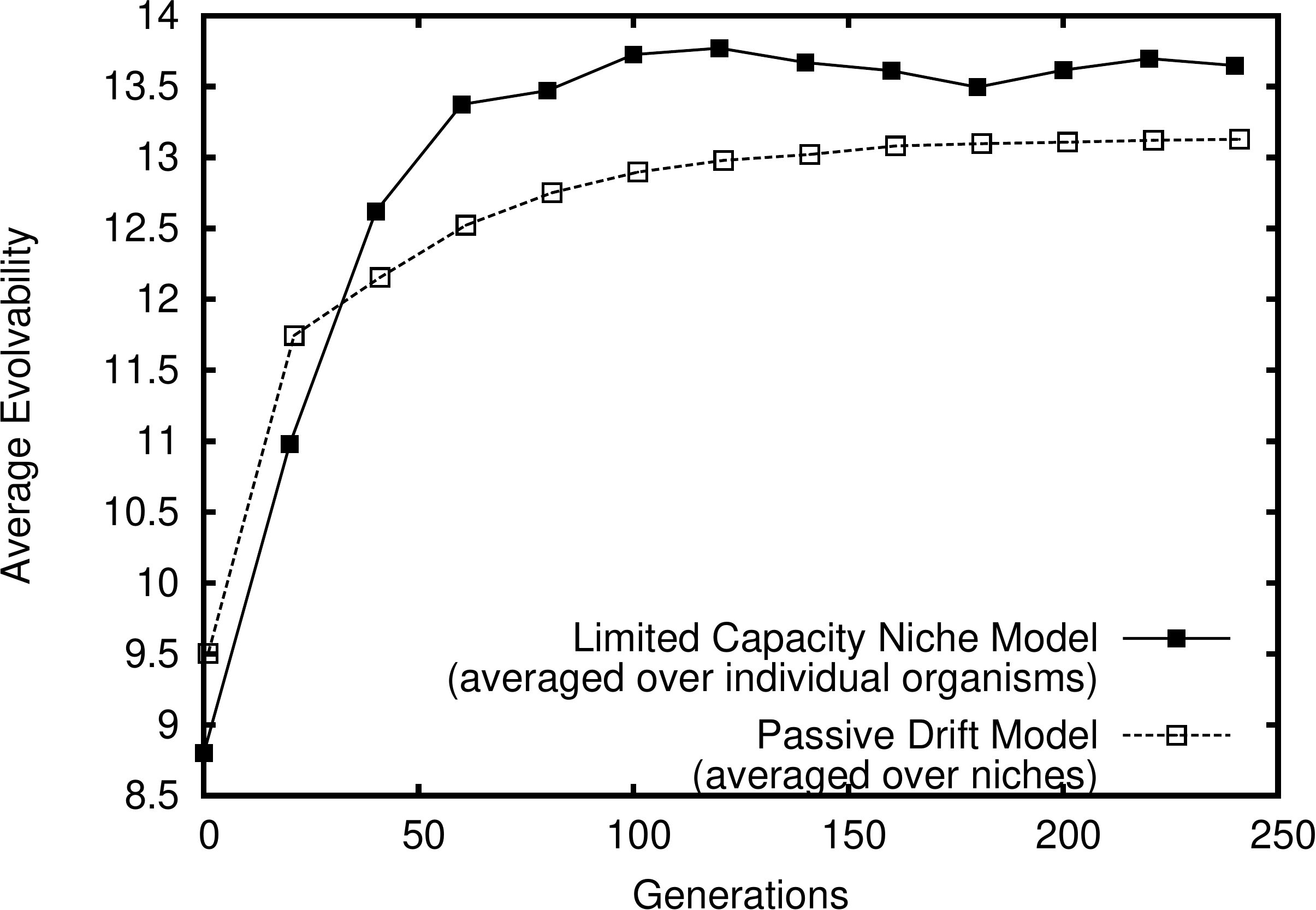}
\caption{\label{fig:model4_evol} \textbf{True increasing evolvability in the evolved robots model with limited capacity niches.} The evolvability of organisms is shown averaged over $50$ independent
simulations that lasted $250$ generations each. It is important to note that because
the population size in the limited-niche model 
is much smaller, \emph{one} generation
of the passive model encompasses more individuals ($2,000,000$) 
than is considered
over \emph{all} generations in the niched model ($1,094,313$ on average).
In other words, the limited-niche model is more directedly and 
more efficiently biased towards evolvability.}
\end{center}
\end{figure}

Like the previous extension to the abstract model, this section 
extends the drifting model with simulated evolved robots to include 
population growth and limited capacity niches.
The idea is to explore whether this more realistic genotype-phenotype
mapping will also exhibit the same 
\emph{accelerated} increase of evolvability seen 
from extending the abstract model. 
The results of this experiment are shown in figure \ref{fig:model4_evol}, and confirm that limiting niche capacity
 also biases population growth more strongly towards increasing evolvability 
in this more concrete genetic space. In other words, the limited niche capacity
models (both the abstract model and the model with evolved robots) demonstrate 
a bias towards true evolvability (i.e.\ when averaged over the entire
population) 
while the passive drift models exhibit a 
weaker bias towards evolvability
that is significant only when averaged over niches.

\subsection*{Practical ER Model with Limited Capacity Niches}

Finally, the last model explores evolvability in a less restricted space of
ANNs. The idea is to examine whether the hypotheses in this paper hold 
even in a commonly used practical ER system. 
In particular, this model explores a limited-capacity niched model (as in the previous two experiments) 
but with a well-established practical neuroevolution method called NEAT \cite{stanley:ec02}. 
Instead of having only three discrete settings for
ANN connection weights (i.e.\ inhibitory, excitatory, or neutral), 
the connection weights in NEAT can vary continuously in strength. 
Additionally, to facilitate increasingly complex evolved
behaviors, the topology of the ANN can itself become increasingly
complex
because of mutations that incrementally introduce new connections and nodes
during evolution. 
As a result of continuous weights and increasing complexity, the space of ANNs 
that NEAT explores is 
effectively infinite and cannot be fully enumerated as in the previous model. 
However, the benefit is
that this class of genetic space is more analogous to that provided by
DNA in nature, 
which is also
open-ended in a similar way.

Because the space cannot be fully enumerated due to computational limits, it is 
impossible to fully characterize the genotypic space and precisely calculate
evolvability (as was done in the fully passive model). 
This full characterization of
the space also facilitated efficient simulation of millions of evolved robots through 
 a precomputed look-up table that mapped a genotype to its niche and evolvability. Such a table is
impossible to construct for the practical ER model, and
thus only the limited niche capacity setup is
implemented here (because it exhibits a driven trend towards evolvability increase
 that is not dependent on a large population size). 
As in the prior robot controller models, the niche space consists of
a discretized grid of the possible locations to which a 
robot can navigate within the
maze. 
The evolvability of a genotype is estimated (because the space cannot be
exhaustively enumerated) by counting how many phenotypic
niches are reachable from many independent random mutations of the
original genotype.
As a control to show that niching robots in a structured way 
is having a positive effect on evolvability,
a comparison experiment is also run where niching is random (i.e.\ an ANN's niche is specified by a 
random number generator instead of being derived from the robot's behavior).
In other words, the random control does not consistently reward behaviors
that are different from those already present in the system.

\begin{figure}
\begin{center}
\includegraphics[height=2in]{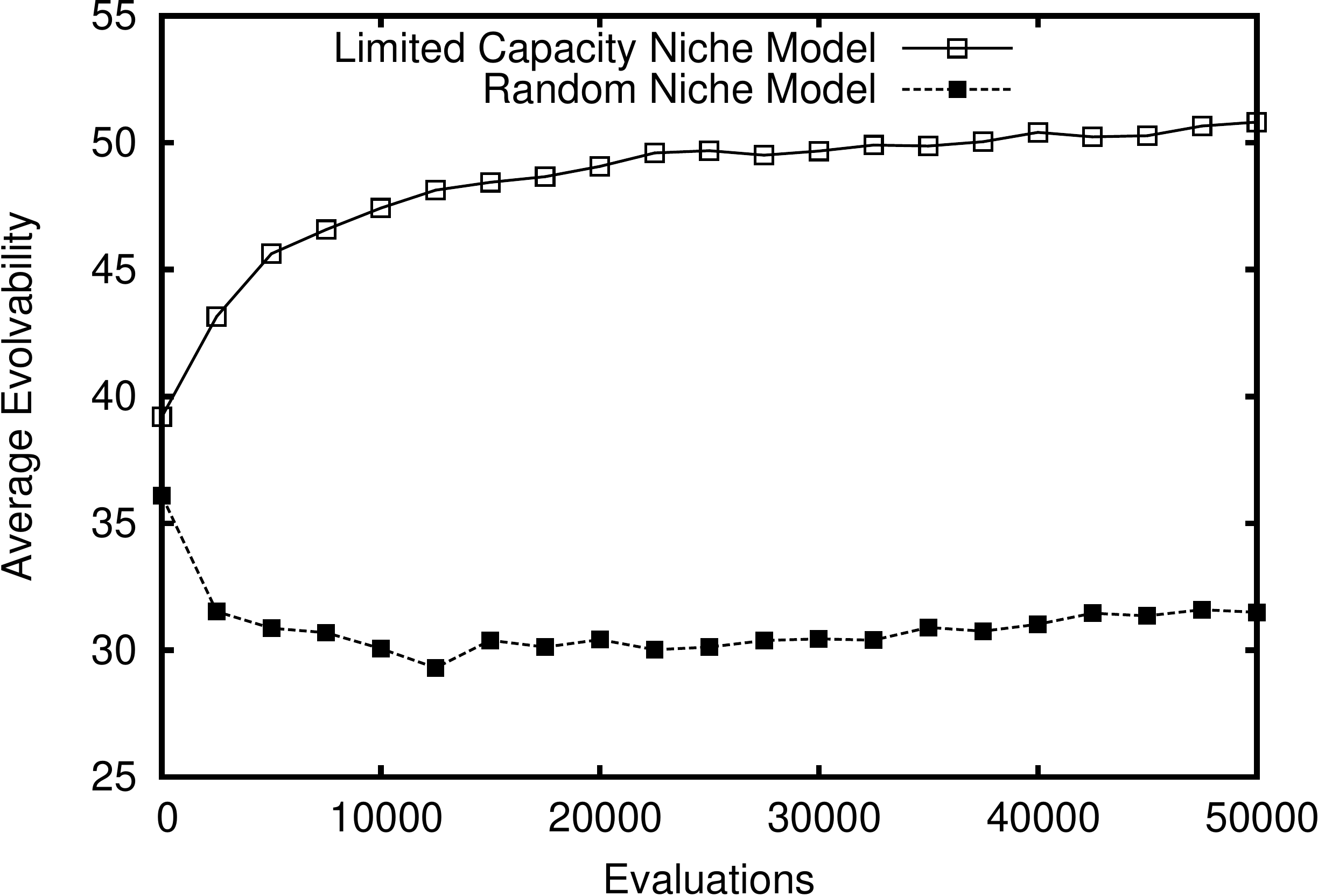}
\caption{\label{fig:model5_evol} \textbf{Evolvability in the practical ER model with limited capacity niches.} The average evolvability of organisms in the population over evolutionary time
is shown, which is itself averaged over $50$ independent simulations that lasted $50,000$ evaluations each. Note that evolvability of an organism is measured as the average number
of different behaviors generated through $300$ random mutations. The main result is that niching based on behavior increases evolvability over a control that randomly
assigns niches independently of an organism's behavior.} 
\end{center}
\end{figure}

The results of these experiments are shown in figure \ref{fig:model5_evol}, which
reinforce the hypotheses in this paper by similarly demonstrating the benefits
for evolvability of population growth with limited niche capacity in a more realistic genetic setting. Note that as in the previous two experiments, limiting
niche capacity encourages true evolvability growth.

\section*{Conclusions}

This paper presented evidence for two non-adaptive explanations for 
the appearance of increasing evolvability over the course of biological
evolution. The first is that an unbiased drifting process over genotypes
can nonetheless produce a distribution of phenotypes (where multiple
individuals in the population may have the same phenotype) biased towards
increasing evolvability, and the
second is that founder effects for discovering niches can provide a
 genotypic bias towards true evolvability increase.
While such non-adaptive explanations do not contradict more
popular adaptive explanations, they call them into question
because the mechanisms shown here require fewer assumptions, i.e.\ they 
result from the structure of the 
genotype-phenotype map and founder effects from uncovering new
niches instead of particular
transient selective pressures. In fact, the results from the
passive drift models suggest even caution in the assumption that 
evolvability in general has truly increased over evolutionary 
time; the superficial \emph{appearance} of
overall increasing evolvability can result
from evolvable organisms filling a larger volume of phenotypic space.

However, even assuming that evolvability has truly increased,
these results still illustrate the danger in habitually 
viewing evolution through the lens of selection pressure. 
An alternative perspective through which to interpret
the results is to view evolution as a
process driven to diversity as it expands through new niches.
Such niche expansion is a ratcheting process, whereby niches
rarely go unfilled after being discovered. 
The founder effect and population growth from uncovering new niches
 serve to bias the genotypic space towards increasing
evolvability because they amplify genomes that diverge phenotypically, which
on average tend to be those that are more evolvable.
Thus if the assumption
of evolvability's heritability holds, then such founder effects in establishing
new niches may yield a persistent
bias towards increasing evolvability -- even in the absence of adaptive
competition between organisms.

In this view 
increasing 
evolvability may simply be an inevitable result of open-ended exploration of
a rich genetic space. Importantly, in nature this passive drive towards
evolvability may have
bootstrapped the evolution of the genotype-phenotype map itself.
That is, the genotypic code and
biological development themselves
are encoded within organisms, and mutations that alter the
structure of the genetic space or genotype-phenotype map
may also lead to more or less
phenotypic possibilities. In this way, the emergence of a complex
evolvable genotypic code and biological development 
may have been bootstrapped from far simpler reproductive processes
by similar non-adaptive mechanisms. In other words, there may be no
selective benefit for development or a complex genetic system, 
 which may do no more than 
potentiate greater phenotypic possibilities. In this way the story of 
biological evolution may be more fundamentally about 
an accelerating drive towards diversity than competition 
over limited resources. 

\section*{Methods}

The following sections provide more details on the experimental models
used in the experiments in this paper.

\section*{Abstract Model Details}

For both abstract models,
at the beginning of the simulation each individual's evolvability 
was initialized to $0.05$. At the beginning of each generation, an 
individual's niche is perturbed with a probability equivalent to its
evolvability, and its evolvability itself
is perturbed with a fixed probability ($0.01$). Changes in evolvability are
drawn from a uniform distribution between $-0.005$ and $0.005$.

In the abstract passive drift model, 
the population consisted of $40,000$ individuals that evolved solely due to
genetic drift for $3,000$ generations. In the model with limited capacity
niches, niches were limited to $5$ individuals each, and the population was
initialized in the first generation with a single individual. Each individual
has two offspring in the next generation, which results in geometric population
growth except when the niche of an offspring is already filled. 
Evolution proceeds for $3,000$ generations.

\section*{Evolved Robot Model Details}

\begin{figure}
\begin{center}
\includegraphics[height=2in]{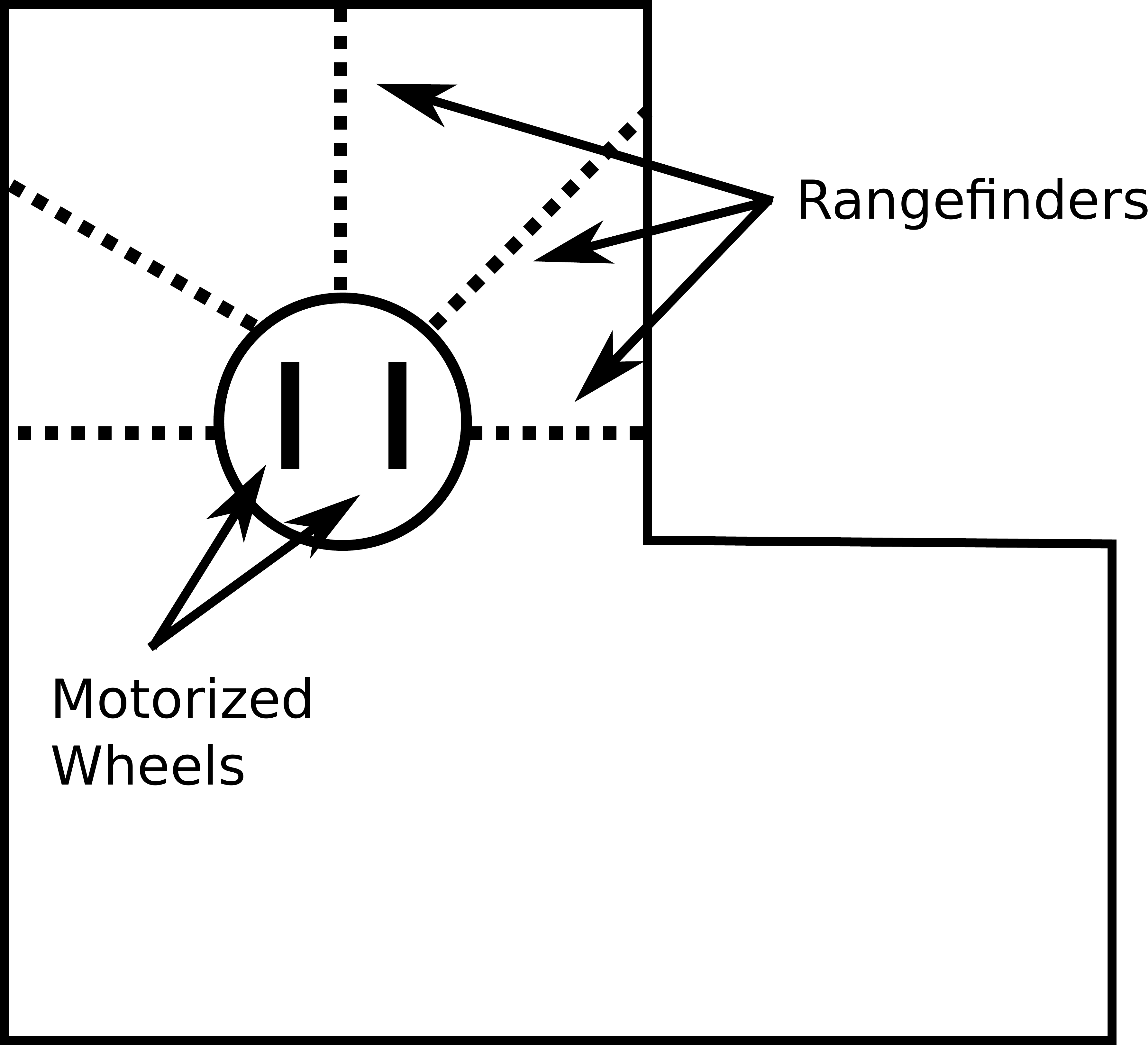}
\caption{\label{fig:robot} \textbf{
Wheeled robot.}
The simulated mobile 
robot is shown that is used in the experiments
with evolved machines. Rangefinder sensors allow the robot to 
perceive obstacles, and the motors controlling its wheels enable the
robot to traverse its environment. 
}
\end{center}
\end{figure}

\begin{figure}
\begin{center}
\includegraphics[height=2in]{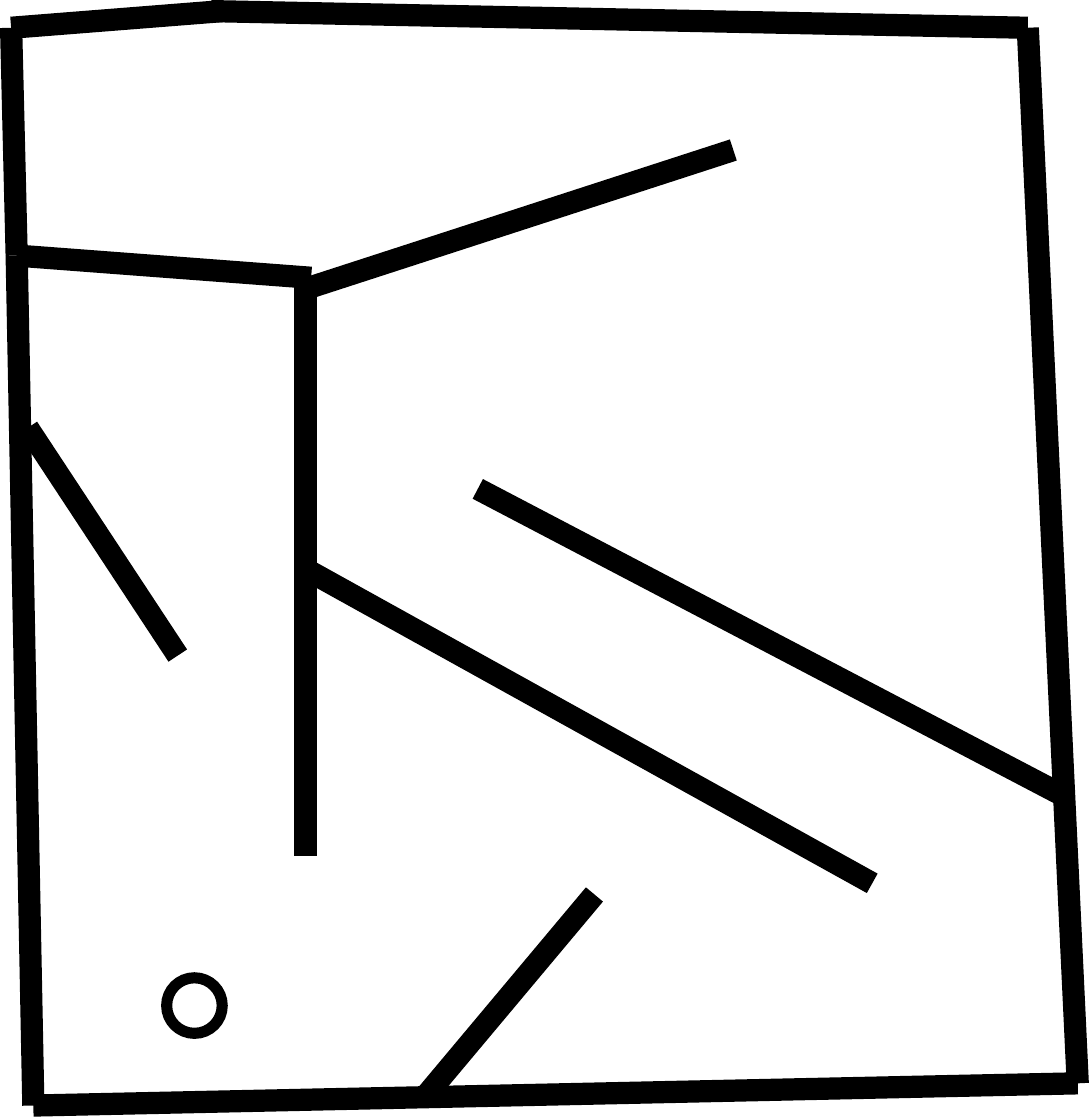}
\caption{\label{fig:hardmaze} \textbf{Maze environment.}
A top-down view of the maze is shown that robots navigate in the experiments with
evolved machines. The circle indicates where a robot begins its trial in
the maze, which lasts for $400$ simulated timesteps. 
}
\end{center}
\end{figure}

In all experiments with evolved machines, a genome that maps to an ANN
 controls a simulated wheeled robot (figure \ref{fig:robot}) with rangefinder sensors in a maze environment (figure \ref{fig:hardmaze}). The experimental
setup follows prior precedent \cite{lehman:ecj11}. 

A uniform $20$x$20$ grid
is superimposed over the maze for calculating
robots' niches. A robot's behavioral
niche (applied for measuring evolvability in both experiments, as well as
to limit population growth in the limited niche capacity experiment) is
determined by
the grid square within which the robot ends at the termination of an evaluation.

\begin{figure}
\begin{center}
\includegraphics[height=2in]{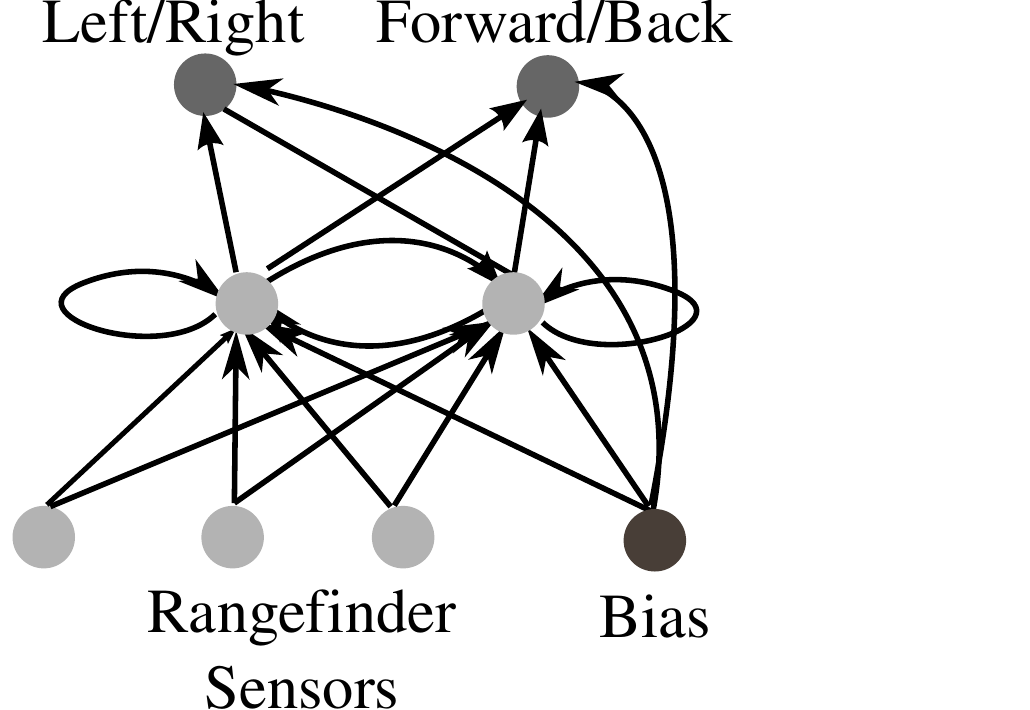}
\caption{\label{fig:ann} \textbf{
Fixed-topology ANN.}
The figure illustrates the fully-connected recurrent ANN with $18$ possible connections that serves as a space of possible controllers for robots embedded in a maze navigation environment. Each connection can either be excitatory (a weight of $1.0$), inhibitory ($-1.0$) or neutral ($0.0$). The activation function in
the ANN is a steepened sigmoid function \cite{stanley:ec02}. The ANN has three rangefinder sensor inputs, two hidden neurons, and two motor outputs.
 }
\end{center}
\end{figure}

The fixed-topology ANN providing the basis for the
enumerated genotypic space of ANNs is shown in figure \ref{fig:ann}.
In particular, the genetic space spans variants of
a fully-connected recurrent ANN with two input nodes, three hidden nodes,
and two output nodes. In total, this kind of ANN has eighteen possible ANN connections that can
either be disabled, excitatory, or inhibatory.
 Thus the space investigated with this
model consists of
$3^{18}$, or $38.7$ million possible genotypes. 
The evolvability of each genome is calculated by first enumerating the genomes  reachable from it by all possible single connection mutations, and then counting
the unique number of behavioral niches that those genomes encode when evaluated in the maze navigation environment.

The drifting model in the enumerated ANN space starts with two
million genotypes initialized in each run to the same random starting genotype. The system
then drifts for $250$ generations.

\section*{Practical ER Model Details}

\begin{figure}
\begin{center}
\includegraphics[height=2in]{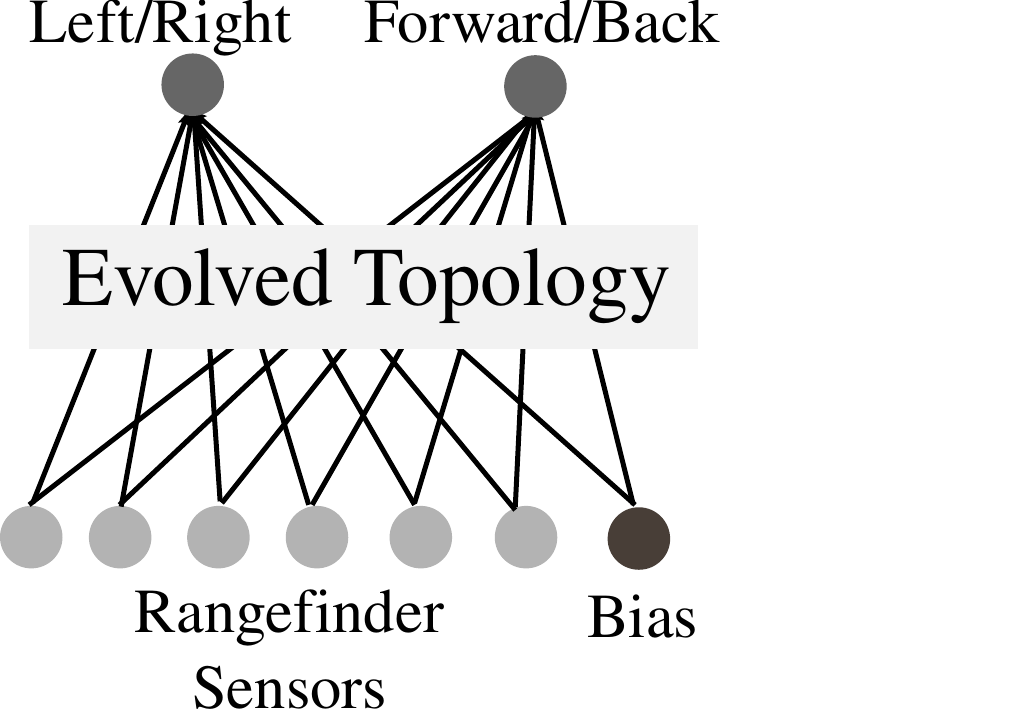}
\caption{\label{fig:ann2} \textbf{
NEAT ANN.}
The initial topology of the ANN in the 
practical ER model is shown. Topologies change during evolution from structural
mutations that add new nodes and connections. In addition, unlike in the
restricted ANN space, connection weights can vary continuously, i.e.\ weight
mutations
perturb connections with values drawn from the uniform distribution, and
weights are capped between $-3.0$ and $3.0$. 
 }
\end{center}
\end{figure}

The practical ER model experiment uses the NEAT algorithm \cite{stanley:ec02},
which relaxes the constraints
of fixed ANN topology and discrete connection weights. The ANN also
provides a greater resolution of sensors, i.e.\ six rangefinder sensors
instead of three. Note that the resolution was reduced in the
fixed topology ANN models for combinatorial reasons. The initial NEAT 
network topology is shown in
figure \ref{fig:ann2}.
All NEAT parameters are the same as those in
Lehman and Stanley \cite{lehman:ecj11}, which has the same experimental
ER maze setup. 

Note that the evolvability of a genome is calculated similarly to how it is
with the enumerated ANN space, i.e.\ by counting the unique number of niches
encoded by genomes in its mutational neighborhood. However, because \emph{all}
reachable mutations cannot be feasibly enumerated in the practical model,
instead $200$ mutations of a given genome 
are randomly sampled to generate a reasonable
\emph{estimate} of its evolvability.

%\section*{References}
% The bibtex filename
\bibliography{nn,evolvability,Novelty,nov2,gptp,lehman}

\begin{thebibliography}{10}
\providecommand{\url}[1]{\texttt{#1}}
\providecommand{\urlprefix}{URL }
\expandafter\ifx\csname urlstyle\endcsname\relax
  \providecommand{\doi}[1]{doi:\discretionary{}{}{}#1}\else
  \providecommand{\doi}{doi:\discretionary{}{}{}\begingroup
  \urlstyle{rm}\Url}\fi
\providecommand{\bibAnnoteFile}[1]{%
  \IfFileExists{#1}{\begin{quotation}\noindent\textsc{Key:} #1\\
  \textsc{Annotation:}\ \input{#1}\end{quotation}}{}}
\providecommand{\bibAnnote}[2]{%
  \begin{quotation}\noindent\textsc{Key:} #1\\
  \textsc{Annotation:}\ #2\end{quotation}}
\providecommand{\eprint}[2][]{\url{#2}}

\bibitem{dawkins:evolution}
Dawkins R (2003) The evolution of evolvability.
\newblock On Growth, Form and Computers : 239--255.
\bibAnnoteFile{dawkins:evolution}

\bibitem{earl:evolvability}
Earl D, Deem M (2004) Evolvability is a selectable trait.
\newblock Proceedings of the National Academy of Sciences of the United States
  of America 101: 11531--11536.
\bibAnnoteFile{earl:evolvability}

\bibitem{kirschner:evolvability}
Kirschner M, Gerhart J (1998) {Evolvability}.
\newblock Proceedings of the National Academy of Sciences of the United States
  of America 95: 8420.
\bibAnnoteFile{kirschner:evolvability}

\bibitem{wagner:evolvability}
Wagner G, Altenberg L (1996) {Complex adaptations and the evolution of
  evolvability}.
\newblock Evolution 50: 967--976.
\bibAnnoteFile{wagner:evolvability}

\bibitem{pigliucci:evolvability}
Pigliucci M (2008) {Is evolvability evolvable?}
\newblock Nature Reviews Genetics 9: 75--82.
\bibAnnoteFile{pigliucci:evolvability}

\bibitem{brookfield:evolvability}
Brookfield J (2001) Evolution: The evolvability enigma.
\newblock Current Biology 11: R106 - R108.
\bibAnnoteFile{brookfield:evolvability}

\bibitem{wagner:robustness}
Wagner A (2008) Robustness and evolvability: a paradox resolved.
\newblock Proceedings of the Royal Society B: Biological Sciences 275: 91--100.
\bibAnnoteFile{wagner:robustness}

\bibitem{bloom:protein}
Bloom J, Labthavikul S, Otey C, Arnold F (2006) Protein stability promotes
  evolvability.
\newblock Proceedings of the National Academy of Sciences 103: 5869--5874.
\bibAnnoteFile{bloom:protein}

\bibitem{dichtel:evolvability}
Dichtel-Danjoy M, F{\'e}lix M (2004) Phenotypic neighborhood and
  micro-evolvability.
\newblock Trends in Genetics 20: 268--276.
\bibAnnoteFile{dichtel:evolvability}

\bibitem{lenski:avida}
Lenski R, Ofria C, Pennock R, Adami C (2003) The evolutionary origin of
  complex.
\newblock Nature .
\bibAnnoteFile{lenski:avida}

\bibitem{nolfi:book00}
Nolfi S, Floreano D (2000) Evolutionary Robotics.
\newblock Cambridge: MIT Press.
\bibAnnoteFile{nolfi:book00}

\bibitem{lehman:ecj11}
Lehman J, Stanley KO (2011) Abandoning objectives: Evolution through the search
  for novelty alone.
\newblock Evol Comp 19: 189-223.
\bibAnnoteFile{lehman:ecj11}

\bibitem{mouret:noveltyjp}
Mouret JB, Doncieux S (2012) Encouraging behavioral diversity in evolutionary
  robotics: an empirical study.
\newblock Evolutionary Computation 20: 91-133.
\bibAnnoteFile{mouret:noveltyjp}

\bibitem{risi:ab10}
Risi S, Hughes C, Stanley K (2010) {Evolving plastic neural networks with
  novelty search}.
\newblock Adaptive Behavior .
\bibAnnoteFile{risi:ab10}

\bibitem{lehman:evolvability}
Lehman J, Stanley KO (2011) Improving evolvability through novelty search and
  self-adaptation.
\newblock In: Evolutionary Computation (CEC), 2011 IEEE Congress on. IEEE, pp.
  2693--2700.
\bibAnnoteFile{lehman:evolvability}

\bibitem{stanley:ec02}
Stanley KO, Miikkulainen R (2002) Evolving neural networks through augmenting
  topologies.
\newblock Evolutionary Computation 10: 99--127.
\bibAnnoteFile{stanley:ec02}

\end{thebibliography}
%\bibliography{template}

%\section*{Figure Legends}
%\begin{figure}[!ht]
%\begin{center}
%%\includegraphics[width=4in]{figure_name.2.eps}
%\end{center}
%\caption{
%{\bf Bold the first sentence.}  Rest of figure 2  caption.  Caption 
%should be left justified, as specified by the options to the caption 
%package.
%}
%\label{Figure_label}
%\end{figure}

%\section*{Tables}
%\begin{table}[!ht]
%\caption{
%\bf{Table title}}
%\begin{tabular}{|c|c|c|}
%table information
%\end{tabular}
%\begin{flushleft}Table caption
%\end{flushleft}
%\label{tab:label}
% \end{table}

\end{document}